# KeypartX: Graph-based Perception (Text) Representation

**Peng Yang**
pengyang@uef.fi

## Abstract

The availability of big data has opened up big opportunities for individuals, businesses and academics to view big into what is happening in their world. Previous works of text representation mostly focused on informativeness from massive words' frequency or cooccurrence. However, big data is a double-edged sword which is big in volume but unstructured in format. The unstructured edge requires specific techniques to transform 'big' into meaningful instead of informative alone.

This study presents KeypartX, a graph-based approach to represent perception (text in general) by key parts of speech. Different from bag-of-words/vector-based machine learning, this technique is human-like learning that could extracts meanings from linguistic (semantic, syntactic and pragmatic) information. Moreover, KeypartX is big-data capable but not hungry, which is even applicable to the minimum unit of text: sentence.

## 1 Introduction

Extracting meaning from text is one of the most important tasks in Natural Language Processing (NLP). Compared to probabilistic techniques (Topic Modelling, Vector Space Model etc.), graphical representation of a text document is more powerful in most of the operations in text such as topology, relational, statistical etc. Moreover, a semantic network is intuitive, natural, and easy to understand [1].

Words/compounds are the basic unit of text corpus to analyse the text representation however, some words (e.g., adjective, verb, noun, and adverb) carry more meaning or information than others[2]. Based on the theoretic concept of 'perception', we propose an Adjective-Verb-Noun network (hereinafter referred to as KeypartX) to search the insight from unstructured text data.

The paper starts with a review on works related to graph-based text representation followed by the introduction of a theoretical concept and implementation of the novel approach. At the end, we draw a conclusion and analyse the technique's methodological features in NLP as well as in real-world application.

## 2 Related Works

In the lately systematic literature review, Eybers & Kahts(2022)[3] identified eighteen text mining techniques(TMT) used to search the insight from unstructured text data. Bag of Words (BoW) and Vector Space Model (VSM) are the fundamental techniques, which convert the textual features to a structured format of vectors. However, BoW-VSM based text mining techniques were widely criticized especially by the school of graph-based text representation[4], since BoW-VSM treats word as independent from other and ignores word-order that result in missing important semantic relations[1].

On the other hand, graphed-based TMTs construct textual network in which nodes denote feature terms and edges denote (co-occurrence, grammatical, semantic or conceptual) relationship between terms[5]. Leskovec et al.(2004)[6] started with deep syntactic analysis of the text and, for each sentence, extract subject–predicate–object in the process of text summarization. With the assistance of VerbNet and WordNet to identify the syntactic/semantic role in a sentence, Hensman (2004) [7]constructed conceptual graphs for English sentences.

Based on a minimum length of a sentence as a unit, Wu et al. (2011)[8] defined a directed semantic graph to represent the text, however the direction only records the ordering information of co-occurrence terms. Different from Wu et al.'s sentential unit, Jin and Srihari (2007)[9] proposed a graph-based text representation, which is capable of capturing term context in documents. The context of a term is defined as the set of its associated concepts within the window boundary. In the same track on application of window boundary, InfraNodus[10] extended the graph-based representation into a large network and applied graph community detection algorithm to cluster different topics in real-world application. However, InfraNodus inclusively connects co-occurrent words either in 2-gram or 4-window instead of labelling any semantic/conceptual relation in the edge.

It is not difficult to figure out that the past works are either limited into the level of sentence or lacks capability of interpretation in 'big text'. Put it linguistically that is semantic and syntactic but not much pragmatic. Most methods start in computer science aiming to increase the informativeness of text representation[11], however informativeness does not necessarily equate to meaningfulness in reality. In order to build up a meaningful representation of text, we shall integrate different forms of linguistic information: semantic (phrase, concept), syntactic(grammar) and pragmatic(interpretation) [12], [13]. Inspired by the psychologic concept of perception, we introduce a novel graph-based text representation-KeypartX which is able to perceive meaningful insights of text data small or big by applying different linguistic features.

## 3 KeypartX

Different from informativeness-oriented tradition, KeypartX is built on the theoretical concept of 'Perception', which prioritizes meaningfulness and capability of interpretation on text representation for a real-world application.

### 3.1 Theoretical Background

Perception has long been an essential concept in philosophy, psychology and business studies. The most common view to perception is a process of attaining awareness or understanding of sensory information[14]. 'Perception' denotes as a form of awareness was the starting point in the fourth Boston Colloquium for the Philosophy of Science[15]. The Aristotelian-Stoic philosophical approach acknowledged that 'perception' is a natural process to "acquire knowledge of an objective world" [16]. Garner et.al (1956)[17]conceived of perception as an "intervening process between stimuli and responses" different from that traditional phycology regarded perception as a discriminatory response. The psychological idea was brought to business study, Solomon et al. (2006, 2017,2019)[18]–[20] stated perception is a process that interprets stimuli into meaning and further constructed the process into three stages: selection, organisation and interpretation. Selection refers to what you want to see, taste, smell, hear and other sensory behaviour. However, based on gestalt psychology principle, people do not perceive a single stimulus in isolation but organize them together to form a general 'feeling'[20]. The third stage of interpretation assigns certain meanings to the organized stimuli.

Perception is composed of three key parts in designative (what or where is perceived), descriptive (how is perceived) and conative (how is re/acted) sections. From a psycholinguistic perspective, nouns (including compound noun, entity) denote whereness and whatness[21][22], adjectives provide descriptive or specific detail[23] and verbs direct sentiment/attitude to nouns[24]. The denotations transform psychological features of perception into linguistic formats. Based on the psycholinguistic transformation, KeypartX constructs an Adjective-Verb-Noun network to represent text using Eq. (1).

a) G = (V, E)

b) V = [ADJ, VERB, NOUN]

c) E = [(ADJ → NOUN), (VERB → NOUN), (NOUN ↔ NOUN, optional)] (1)

Where → denotes semantic relation and ↔ is co-occurrence of two nodes. The nodes of adjective, verb, noun roughly represent descriptive, conative and conative parts respectively. Adjective to noun and verb to noun are single-directional edges (semantic relations) and bi-directional edges(co-occurrence) of noun-noun will be also connected (N-N edges are optional). Adjective-Verb→Noun network transforms psychologic perception into a linguistic network. Perception refers to what we understand the meaning of world and the network is composed by keys parts of perception in textual form. Theoretically, KeypartX affords a feasible method to graph the insights from unstructured text.

### 3.2 Methodology

KeypartX generates text representation through two main phases: Natural Language Processing and Perceptual Network Analysis. The methodology to construct KeypartX is illustrated in Figure 1.

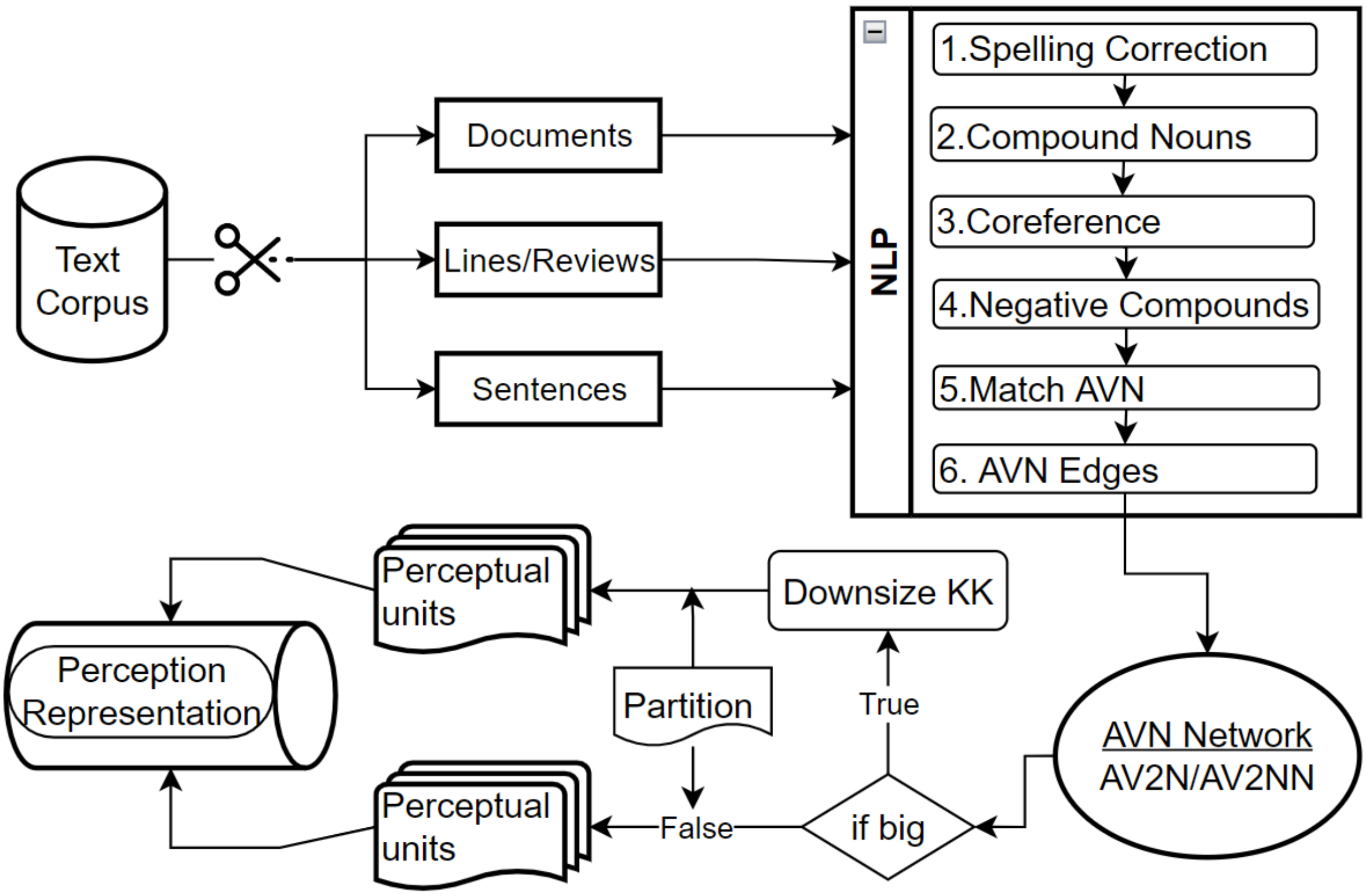

Figure 1: KeypartX Architecture

### 3.2.1 Natural Language Processing

In the first phase, text corpus could be segmented by document, lines/reviews or further into sentences level. Here we created a text to exemplify the steps of NLP.

Example Text:

```
Thai food was great,delicousr and not expensive, we loved it. We visited 3 beach resorts
, they are higly recommened... We had "Fire-Vodka" !!!
```

The text especially online content is often written unchecked, then spelling correction goes first. The second step is to find compound nouns which are more descriptive than separate nouns. Compound nouns are composed of quoted and hyphen-connected words, entity (such as “New York”) and noun-noun compounds(optional). The third step aims to replace the pronoun with noun by coreference so that would increase noun’s representation. With respect to adjective and verb, negative compounds are necessary to delineate the nouns more exactly e.g., ‘not good coffee’ should be extracted as ‘not-good’ instead of ‘good’ alone. In the fifth step, we applied the conception-based linguistic patterns to match Adjective, Verb and Noun. Finally, a set of edges were created: Adjective→Noun, Verb→Noun and Noun ↔Noun (optional but recommended). Figure 2 shows the final target of NLP of example text.

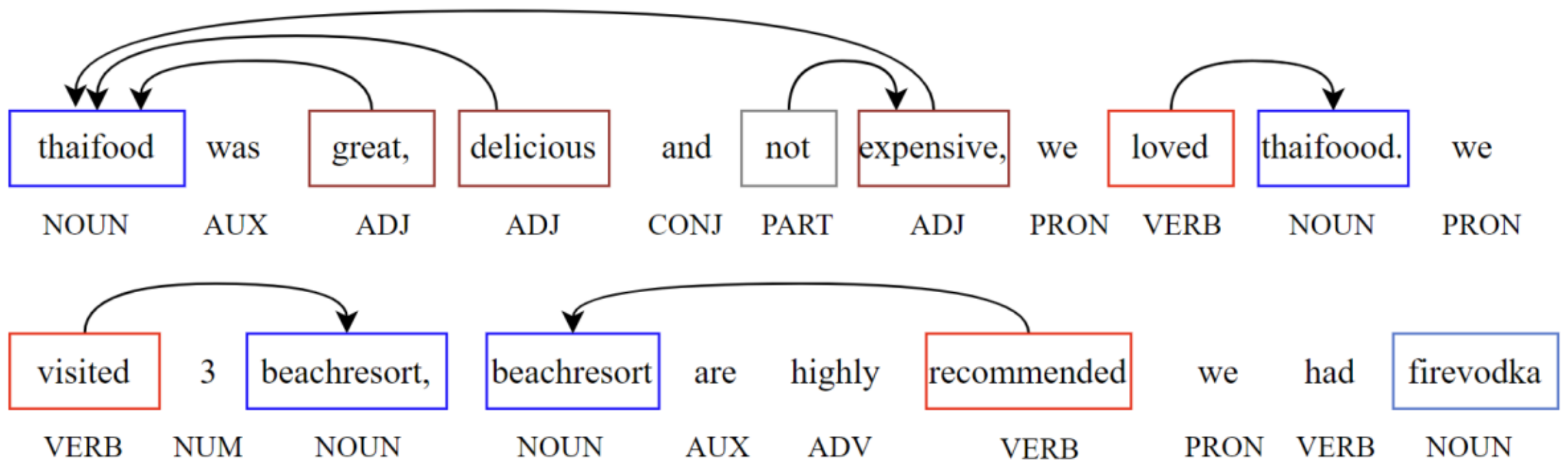

Figure 2: NLP Target

Step 1: Spelling Correction (delicousr/higly/ recommened)

```
Thai food was great,delicious and not expensive, we loved it. We visited 3 beach resorts
, they are highly recommended... We had "Fire-Vodka" !!!
```

Step 2 : Compound Nouns

```
thaifood was great, delicious and not expensive, we loved it. we visited 3 beachresort,
they are highly recommended… we had firevodka
```

Step 3: Coreference

```
thaifood was great, delicious and not expensive, we loved thaifood. we visited 3
beachresort, beachresort are highly recommended… we had firevodka
```

Step 4: Negative Compounds

If any, negative words are compounded with adjective or verbs to create a new more rigorous representation. Negative words used in this case are such like 'hardly', 'scarcely', 'barely', 'no', 'not', 'none', 'neither', 'nor', 'never'.

```
thaifood was great, delicious and notexpensive, we loved thaifood. we visited 3
beachresort, beachresort are highly recommended… we had firevodka
```

Step5: Match AV2N

There are four patterns to match Adjective/Verb to Noun, in which copular verbs (e.g., be, get, taste, smell, seems etc.) and passive voice verbs (e.g., be and get) were used to form patterns of Noun-Copular-Adjective and Noun-Passive-Verb respectively.

```
pattern1 = [{"POS": "NOUN","OP":"+"},{"LOWER": "Copular","OP":"+"},{"POS": "ADJ","OP":"+
"}]matcher.add('N+A', [pattern1])

pattern2 = [{"POS": "ADJ","OP":"+"},{"POS": "NOUN","OP":"+"}]
matcher.add('A+N', [pattern2])

pattern3 = [{"POS": "VERB"},{"POS": "NOUN","OP":"+"}]
matcher.add('V+N', [pattern3])

pattern4 = [{"POS": "NOUN","OP":"+"},{"LOWER": "Passive","OP":"+"},{"POS": "VERB"}]
matcher.add('N+V', [pattern4])
```

In this step, all adjectives, verbs and nouns are mapped with new ones by adding 2a,2v,2n respectively, due to some words have two or more parts of speech such as ‘love’ could be both noun and verb.

```
thaifood2n be great2a, delicious2a notexpensive2a , love2v thaifood2n .
beachresort2n , beachresort2n be recommend2v . . .  firevodka2n
```

Step 6: AV2N or AV2NN Edges

```
AV2N_edges = [['great2a', 'thai2nfood2n'], ['delicious2a', 'thaifood2n'],
['notexpensive2a', 'thaifood2n'], ['love2v', 'thai2nfood2n'], ['recommend2v',
'beachresort2n']]

if noun-noun:
    [('thai2nfood2n', 'firevodka2n'), ('thaifood2n', 'beachresort2n'),
     ('firevodka2n’,'beachresort2n')]
then:
    AV2NN edges = AV2N edges + NN edges
```

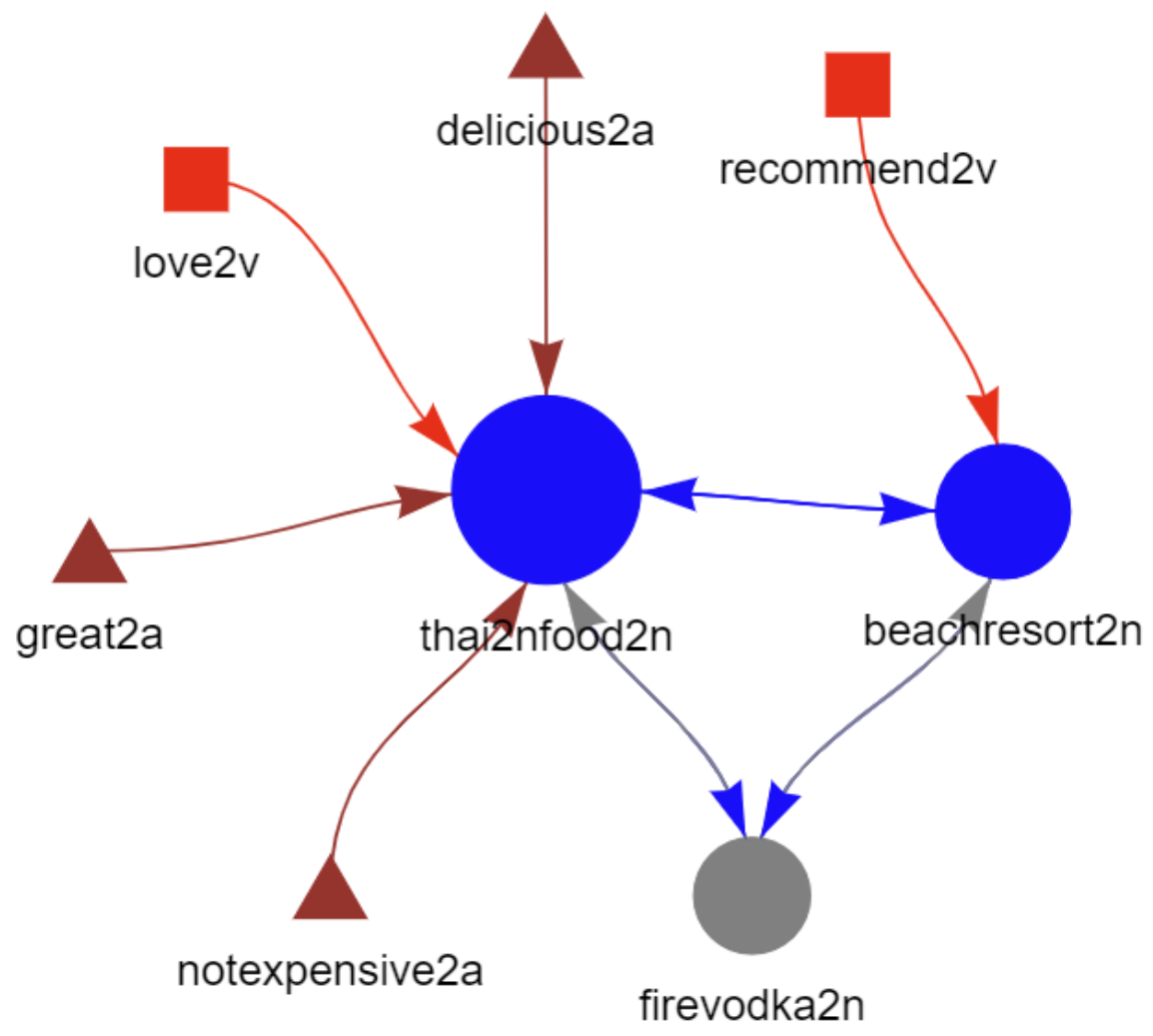

Figure 3: Perceptual Unit (Greedy NN)

The AV2NN network was constructed based on AV2N and NN edges. The AV2NN performs better than AV2N. To be noted: NN edges are either only nouns in AV2N or all nouns (greedy) in text such as firevodka2n in this case (Gray node in Figure 3).

If text data are so big that network needs to be further downsized and partitioned. In the second phase, perceptual units are generated and then ensembled into a general perception if necessary.

### 3.2.2 Perceptual Network Analysis

In the second phase of perceptual network analysis, we used a corpus of 118 reviews posted on TripAdvisor by those who visited Thailand[1].

It is to be noted that only the following 'conative verbs' were kept in this case, since not all verbs are meaningful such as 'go', 'come' and 'have' and etc.

```
con_verbs = ['enjoy','love','like','adore','avoid','revisit','desire','dislike','hate','
wish','hope','appreciate','value','recommend','unrecommend','astonish','impress','please
','satisfy','unsatisfy','surprise','mean','mind']
```

Following the first phase of NLP, An AV2NN network was created comprising of 697 nodes (232 adjectives, 10 verbs and 455 nouns) and 4546 edges (900 AV2N and 3646NN edges).

The network was downsized by setting k-weight as 2 (k is the size of edge weight) and k-core as 2(the size k of node degree, k=2 in this case) resulting in 145 nodes and 198 edges. The downsized network (Figure 4a) generated some unconnected nodes which are further removed in the step of partition.

We implemented community detection in directed graph proposed by Leicht and Newman (2008)[25] (Eq.2, implemented in Traag's leidenalg 0.8.10[26]) to partition the network. Finally, it is partitioned into 16 communities including three without adjective or verb (Figure 4b).

$$Q = \sum_{ij}\left(A_{ij} - \gamma \frac{k_i^{out}k_j^{in}}{m}\right)\delta(\sigma_i, \sigma_j) \qquad (2)$$

Where A is the adjacency matrix, $k_i^{out}$ and $k_j^{in}$ refers to respectively the outdegree and indegree of node i, and $A_{ij}$ refers to an edge from i to j, $\sigma_i$ denotes the community of node I and $\delta(\sigma_i, \sigma_j)$ = 1 if $\sigma_i$=$\sigma_j$ and 0 otherwise, m is the total number of edges in the network.

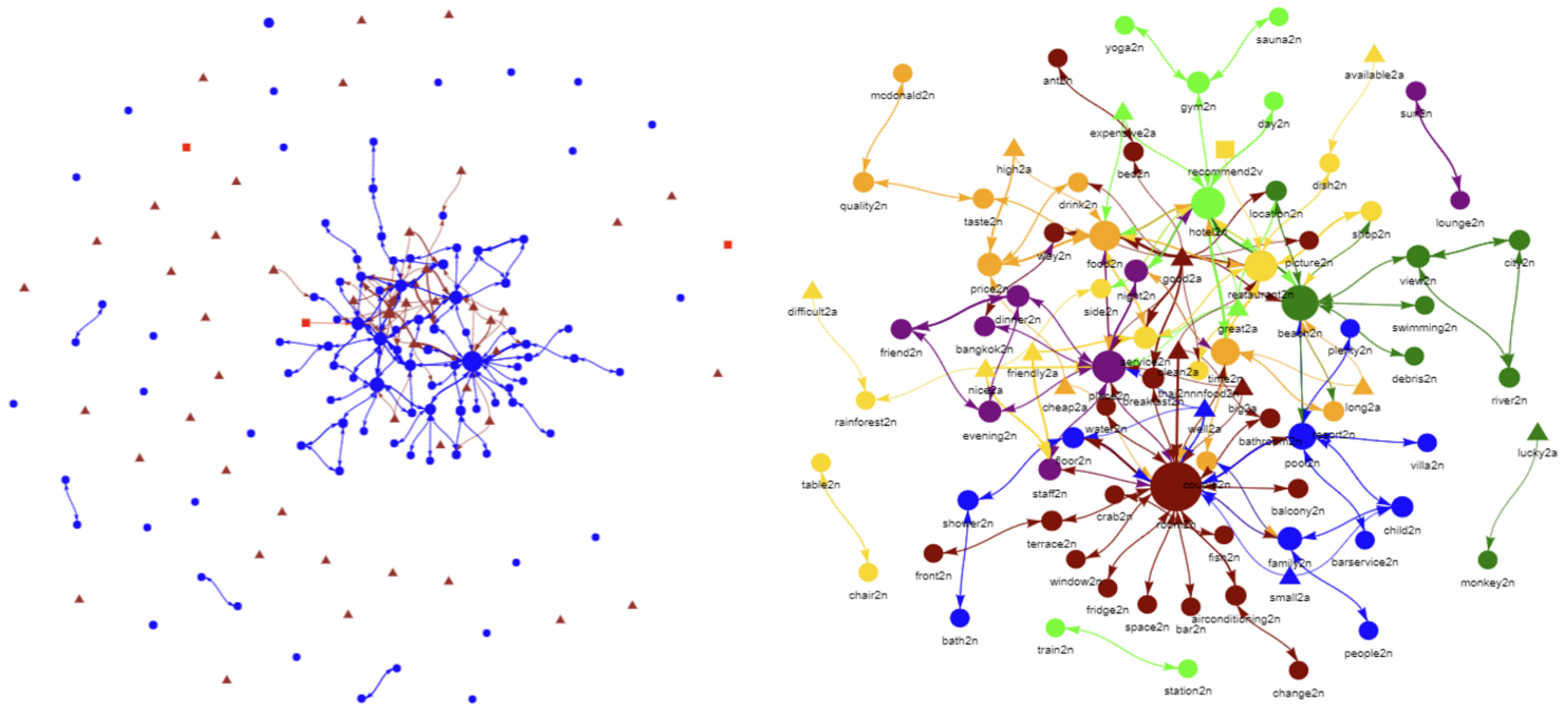

*(a)* *(b)*

Figure 4: Conceptual Network Analysis to downsize (**a**) and partition(**b**)the network.

---

[1] https://github.com/pengKiina/KeypartX/blob/main/Thailand_text.csv

In general, partitioned network is still too complicated to generate perceptions. Moreover, a variety of nodes in a specific community are also collected with nodes in other community. It would downgrade perception representation of community if shared nodes were ruled out. In order to address the two deficiencies, a set of special communities i.e., perceptual units with grey nodes (nodes are shared by other community) were selected out the network. The perceptual units are manageable parts of the whole network to extract the meaning of the text representation.

Figure 5 shows the first two perceptual units in this case. The units display the size and colour of both nodes and edges, in which the blue circle, red square and brown triangle represent noun, verb and adjective respectively.

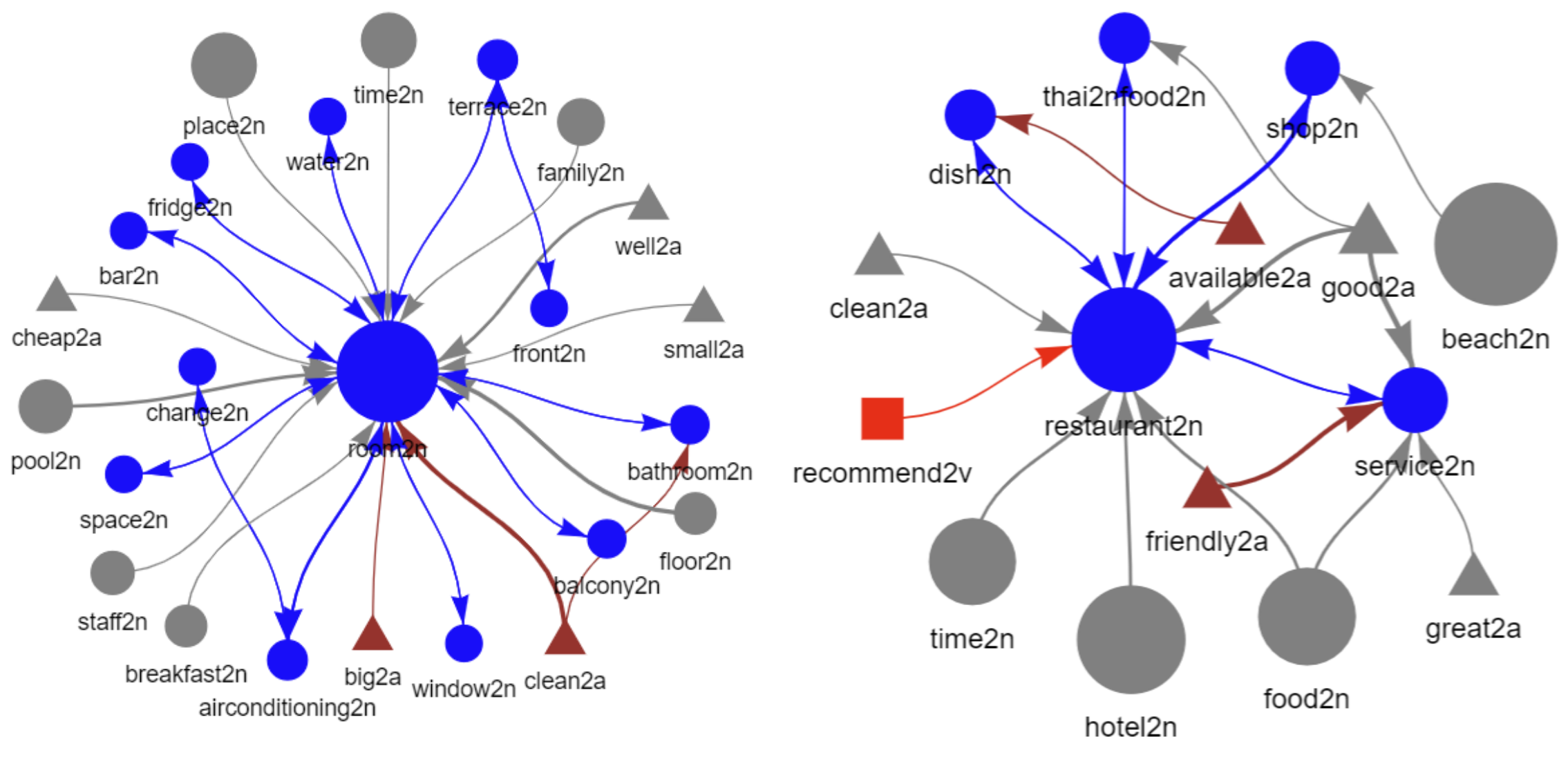

Figure 5: Grey plus perceptual units, **Left** and **Right** units are graphed based on words in community 1 and community 2.

The grey nodes are words not in community 1 and 2 but necessary to reinforce the representation of perception.

```
[community 1] clean2a, room2n, big2a, bathroom2n, airconditioning2n,
     space2n, change2n, bar2n, fridge2n, window2n, balcony2n,
     front2n, terrace2n, water2n
[community 2] service2n, friendly2a, restaurant2nnn, recommend2v,
     thai2nfood2n(compound), available2a, dish2n, shop2n
```

The communities in network, to some extent, is similar to topics in machine learning techniques such as topic modelling, however the latter lacks readability without semantic relations. MLs present each topic in a set of key words(nodes) but missing edges to cohere them in a meaningful logic. In addition to the readable feature, KeypartX represents unstructured text in a more meaningful and human-like linguistic network. In the unit network, it is easy to find the importance of nodes (size by degree) and semantic relations (edges width by weight). The key perceptions in grey units (Figure 5) can easily be discerned in dynamic network or transformed to tabular form (Table1).

| Key Perception (Degree) | Adjective (weight) | Verb (weight) | Noun (weight) |
|---|---|---|---|
| **Room (41)** | Clean (7), Big (2), Cheap (2), Small (2), etc. | | Window (2), Fridge (2), Water (2), Bathroom (2), etc. |
| **Restaurant (18)** | Good (5), Friendly (5), Clean (2) | Recommend (2) | Food (3), Thai food (2), Service (2), Dish (2), etc. |

Table 1: Semantic Relations of Key Perceptions

Different from ML techniques, a manual validation was conducted on the text representation by reading through all 118 reviews. The results represent the main idea of text that customers liked hospitality service and enjoyed beach-night life in Thailand.

## 4 Conclusion

We presented a novel approach of perception (text in general) representation by generating semantic network from unstructured text data and applying grey plus perceptual unit to optimize its interpretation. Compared to informativeness-oriented methods, KeypartX represents text in a more meaningful and interpretable graphs regardless of how big or small the data. KeypartX is built on theoretic concept of perception that in turn, opens a new methodological technique for studies in philosophy, psychology and business. In big data era, businesses are especially frustrated to dig the insight of what consumers perceive on their product or service from massive sets of data. Nevertheless, KeypartX still needs improvement such as it does not specify the degree to which a noun is perceived e.g., ‘like Thailand’ is less informative than ‘like Thailand very much’. The work is ongoing and efforts are continuing to improve the application in real life.

## Reference

[1] T. L. Pa, M. Kumari, T. Singh, and M. Ahsan, 'Semantic Representations in Text Data', *IJGDC*, vol. 11, no. 9, pp. 65–80, Sep. 2018, doi: 10.14257/ijgdc.2018.11.9.06.

[2] G. A. Miller, 'WordNet: a lexical database for English', *Commun. ACM*, vol. 38, no. 11, pp. 39–41, Nov. 1995, doi: 10.1145/219717.219748.

[3] S. Eybers and H. Kahts, 'In Search of Insight from Unstructured Text Data: Towards an Identification of Text Mining Techniques', in *Digital Science*, Cham, 2022, pp. 591–603. doi: 10.1007/978-3-030-93677-8_52.

[4] A. H. Osman and O. M. Barukub, 'Graph-Based Text Representation and Matching: A Review of the State of the Art and Future Challenges', *IEEE Access*, vol. 8, pp. 87562–87583, 2020, doi: 10.1109/ACCESS.2020.2993191.

[5] S. S. Sonawane and D. P. A. Kulkarni, 'Graph based Representation and Analysis of Text Document: A Survey of Techniques'. International Journal of Computer Applications, 2014.

[6] J. Leskovec, M. Grobelnik, and N. Milic-Frayling, 'Learning Semantic Graph Mapping for Document Summarization', Jan. 2004.

[7] S. Hensman, 'Construction of Conceptual Graph Representation of Texts', in *Proceedings of the Student Research Workshop at HLT-NAACL 2004*, Boston, Massachusetts, USA, May 2004, pp. 49–54. Accessed: Sep. 08, 2022. [Online]. Available: https://aclanthology.org/N04-2009

[8] J. Wu, Z. Xuan, and D. Pan, 'Enhancing text representation for classification tasks with semantic graph structures', *International Journal of Innovative Computing, Information and Control*, vol. 7, May 2011.

[9] W. Jin and R. K. Srihari, 'Graph-based text representation and knowledge discovery', in *Proceedings of the 2007 ACM symposium on Applied computing*, New York, NY, USA, Mar. 2007, pp. 807–811. doi: 10.1145/1244002.1244182.

[10] D. Paranyushkin, 'InfraNodus: Generating Insight Using Text Network Analysis', in *The World Wide Web Conference*, New York, NY, USA, May 2019, pp. 3584–3589. doi: 10.1145/3308558.3314123.

[11] Z. Wu and C. L. Giles, 'Measuring Term Informativeness in Context', in *Proceedings of the 2013 Conference of the North American Chapter of the Association for Computational Linguistics: Human Language Technologies*, Atlanta, Georgia, Jun. 2013, pp. 259–269. Accessed: Sep. 07, 2022. [Online]. Available: https://aclanthology.org/N13-1026

[12] C. A. Gurr, 'Effective Diagrammatic Communication: Syntactic, Semantic and Pragmatic Issues', *Journal of Visual Languages & Computing*, vol. 10, no. 4, pp. 317–342, Aug. 1999, doi: 10.1006/jvlc.1999.0130.

[13] G. R. Kuperberg *et al.*, 'Common and Distinct Neural Substrates for Pragmatic, Semantic, and Syntactic Processing of Spoken Sentences: An fMRI Study', *Journal of Cognitive Neuroscience*, vol. 12, no. 2, pp. 321–341, Mar. 2000, doi: 10.1162/089892900562138.

[14] O. Qiong, 'A Brief Introduction to Perception', *Studies in Literature and Language*, vol. 15, no. 4, Art. no. 4, Oct. 2017, doi: 10.3968/10055.

[15] R. Efron, 'What is Perception?', in *Proceedings of the Boston Colloquium for the Philosophy of Science 1966/1968*, vol. 4, R. S. Cohen and M. W. Wartofsky, Eds. Dordrecht: Springer Netherlands, 1969, pp. 137–173. doi: 10.1007/978-94-010-3378-7_4.

[16] B. Maund, *Perception*. Chesham, Bucks: Acumen, 2003.

[17] W. R. Garner, H. W. Hake, and C. W. Eriksen, 'Operationism and the concept of perception.', *Psychological Review*, vol. 63, no. 3, pp. 149–159, 1956, doi: 10.1037/h0042992.

[18] M. R. Solomon, Ed., *Consumer behaviour: a European perspective*, 3rd ed. Harlow, England ; New York: Financial Times/Prentice Hall, 2006.

[19] M. R. Solomon, *Consumer behavior: buying, having, and being*, Twelfth Edition. Boston: Pearson, 2017.

[20] M. R. Solomon, *Consumer behaviour: a European perspective*, Seventh edition. Harlow: Pearson Education, 2019.

[21] A. Wierzbicka, ‘What’s in a Noun? (Or: How Do Nouns Differ in Meaning from Adjectives?)’, *Studies in Language*, vol. 10, Jan. 1986, doi: 10.1075/sl.10.2.05wie.

[22] M. Baroni and R. Zamparelli, ‘Nouns are Vectors, Adjectives are Matrices: Representing Adjective-Noun Constructions in Semantic Space’, in *Proceedings of the 2010 Conference on Empirical Methods in Natural Language Processing*, Cambridge, MA, Oct. 2010, pp. 1183–1193. Accessed: Jul. 20, 2022. [Online]. Available: https://aclanthology.org/D10-1115

[23] A. Fowler, *The Little Brown Handbook, 11th Edition*, 11th edition. Boston: Pearson, 2011.

[24] M. Karamibekr and A. Ghorbani, ‘Verb Oriented Sentiment Classification’, Dec. 2012, pp. 327–331. doi: 10.1109/WI-IAT.2012.122.

[25] E. A. Leicht and M. E. J. Newman, ‘Community Structure in Directed Networks’, *Phys. Rev. Lett.*, vol. 100, no. 11, p. 118703, Mar. 2008, doi: 10.1103/PhysRevLett.100.118703.

[26] V. A. Traag, ‘leidenalg: Leiden is a general algorithm for methods of community detection in large networks.’ 2022. Accessed: Sep. 15, 2022. [MacOS :: MacOS X, Microsoft :: Windows, POSIX]. Available: https://github.com/vtraag/leidenalg